\documentclass[runningheads]{llncs}
\usepackage[T1]{fontenc}
\usepackage{graphicx}
\usepackage{subcaption}
\usepackage{amsmath}
\usepackage{amssymb}
\usepackage{svg}
\usepackage{booktabs} 
\usepackage{pifont} 
\usepackage{pgfplots} 
\usepackage{wrapfig} 
\usepackage{array} 
\usepackage[colorlinks=true, allcolors=blue]{hyperref} 
\usepackage{multirow}
\usepackage{tabularx}
\newcommand{\cmark}{\textcolor{green}{\ding{51}}} 
\newcommand{\xmark}{\textcolor{red}{\ding{55}}} 
\newcommand{\samplecount}{852} 

\begin{document}
\title{BikeActions: An Open Platform and Benchmark for Cyclist-Centric VRU Action Recognition}
\titlerunning{BikeActions}  
%
\author{
Max A. Buettner\inst{1,2}\orcidID{0009-0006-7939-6256} \and
Kanak Mazumder\inst{1,2}\orcidID{0009-0006-6806-8388} \and
Luca Koecher\inst{1} \and
Mario Finkbeiner\inst{1}\orcidID{0009-0009-9510-4534} \and
Sebastian Niebler\inst{1} \and
Fabian B. Flohr\inst{1}\orcidID{0000-0002-1499-3790}
}

\authorrunning{M.~A.~Buettner et al.}
\institute{
Munich University of Applied Sciences, Intelligent Vehicles Lab (IVL),\\
Munich, Germany\\
\email{\{max.buettner, kanak.mazumder, koecher, mario.finkbeiner, sebastian.niebler, fabian.flohr\}@hm.edu}
\and 
\inst{†}Authors contributed equally.
}

\maketitle              

\begin{abstract} 
Anticipating the intentions of Vulnerable Road Users (VRUs) is a critical challenge for safe autonomous driving (AD) and mobile robotics. While current research predominantly focuses on pedestrian crossing behaviors from a vehicle's perspective, interactions within dense shared spaces remain underexplored. To bridge this gap, we introduce \textit{FUSE-Bike}, the first fully open perception platform of its kind. Equipped with two LiDARs, a camera, and GNSS, it facilitates high-fidelity, close-range data capture directly from a cyclist's viewpoint. Leveraging this platform, we present \textit{BikeActions}, a novel multi-modal dataset comprising \samplecount{} annotated samples across 5 distinct action classes, specifically tailored to improve VRU behavior modeling. We establish a rigorous benchmark by evaluating state-of-the-art graph convolution and transformer-based models on our publicly released data splits, establishing the first performance baselines for this challenging task. We release the full dataset together with data curation tools, the open hardware design, and the benchmark code to foster future research in VRU action understanding under \href{https://iv.ee.hm.edu/bikeactions/}{\textit{BikeActions}}. 
\end{abstract}

\section{Introduction}
\label{sec:introduction}

Safe and scalable autonomous systems, whether self-driving cars or mobile service robots, require a comprehensive perception of the environment and the traffic participants operating within it. In urban settings, this challenge is most acute in shared spaces, where autonomous agents must interact closely with VRUs such as pedestrians and cyclists. These groups are disproportionately affected by traffic risks~\cite{who}, and their movements are often sudden, entangled, and driven by subtle non-verbal cues. Gestures, head orientation, and body posture convey critical intent that current robotic and automotive perception systems struggle to interpret.

While modern perception research predominantly operates in a data-driven manner, it remains fundamentally constrained by the data it is trained on. The vast majority of large-scale datasets prioritize 3D object detection, tracking, and trajectory prediction, often treating VRUs as rigid bounding boxes. This abstraction undervalues the complexity of human behavior required for safe navigation in crowded shared spaces. Although some pipelines utilize 3D pose history for motion prediction, they lack the fine-grained action labels necessary to learn the causal link between body pose and intent. Without an explicit supervision, models often fail to decode the rich signals encoded in human motion, such as a cyclist’s hand signal before a turn or a pedestrian’s tentative posture at a crossing, limiting the deployment of robots and vehicles in human-centric environments.

Existing datasets are insufficient to address this challenge, as they lack the specific data required for fine-grained, close-range VRU action recognition. For example, large-scale benchmarks like Kinetics~\cite{kinetics} provide a wealth of general human actions but are curated from web videos, lacking the specific context, multimodal sensor streams, and consistent viewpoints of urban driving. Conversely, while recent automotive datasets like ROAD-Waymo~\cite{road-waymo-github} offer action labels on top of synchronized sensor data, they remain inherently limited by their vehicular perspective. Captured from a high-elevation, car-centric viewpoint, these datasets observe VRUs from a distance and fail to capture the intimate, close-range interactions and unique perspective of a cyclist navigating shared spaces. This leaves a critical gap for a high-fidelity, multimodal benchmark captured from a true VRU-centric viewpoint.

\begin{figure*}[!t] 
    \centering

    
    \begin{subfigure}{0.32\linewidth}
        \includegraphics[width=\linewidth]{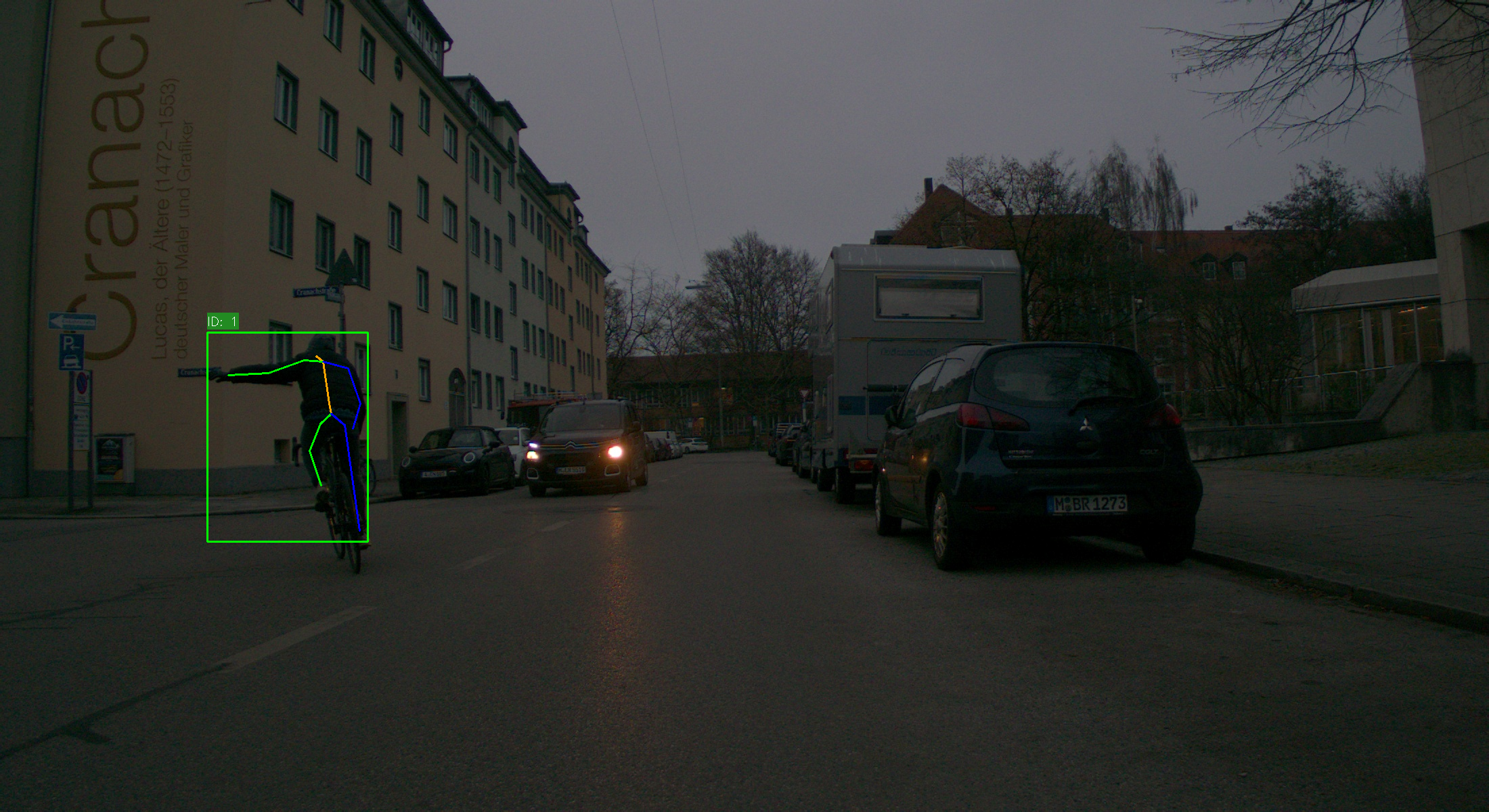} 
        \label{fig:glimpse-rgb1}
    \end{subfigure}
    \hfill 
    \begin{subfigure}{0.32\linewidth}
        \includegraphics[width=\linewidth]{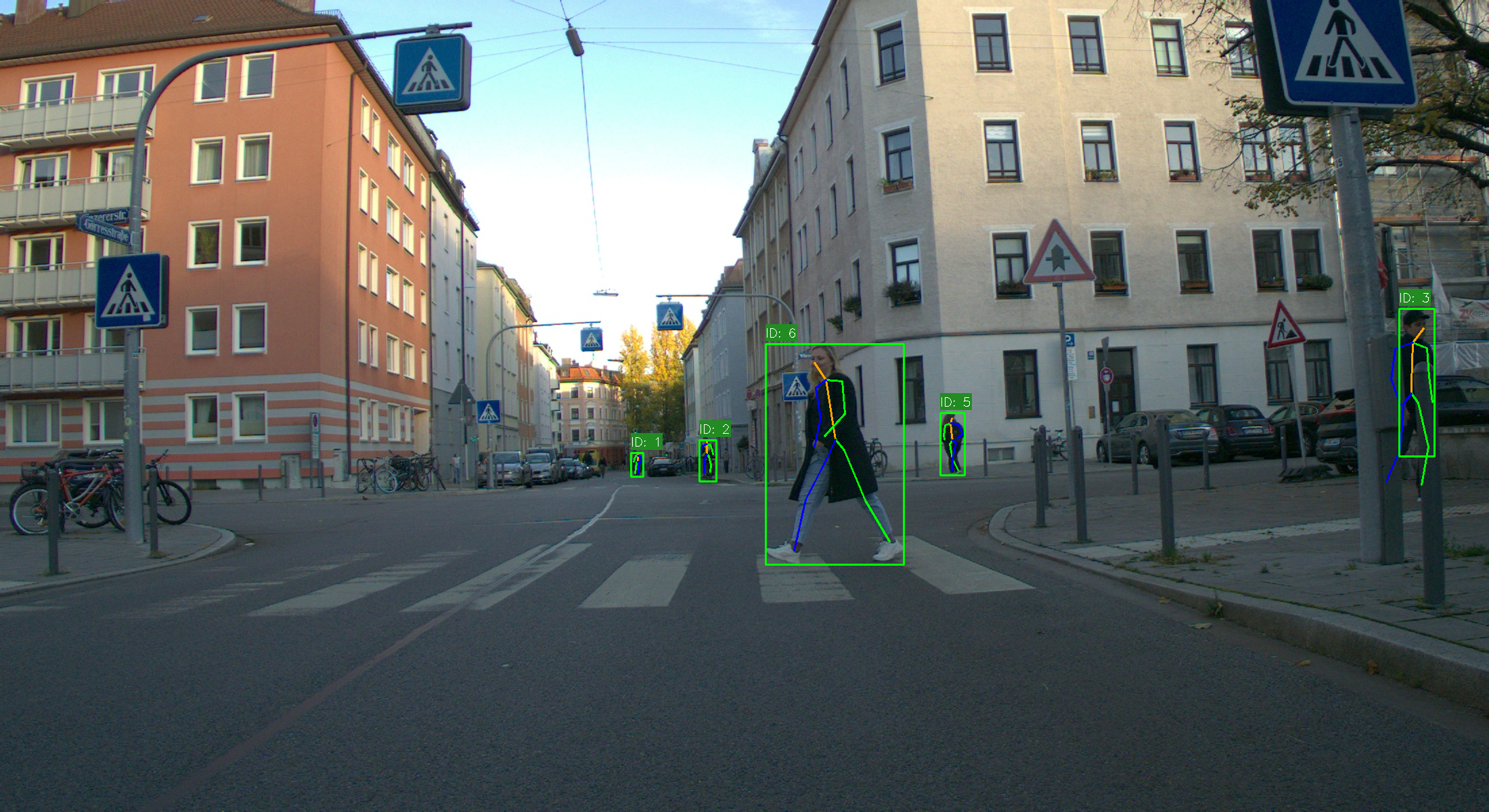} 
        \label{fig:glimpse-rgb2}
    \end{subfigure}
    \hfill
    \begin{subfigure}{0.32\linewidth}
        \includegraphics[width=\linewidth]{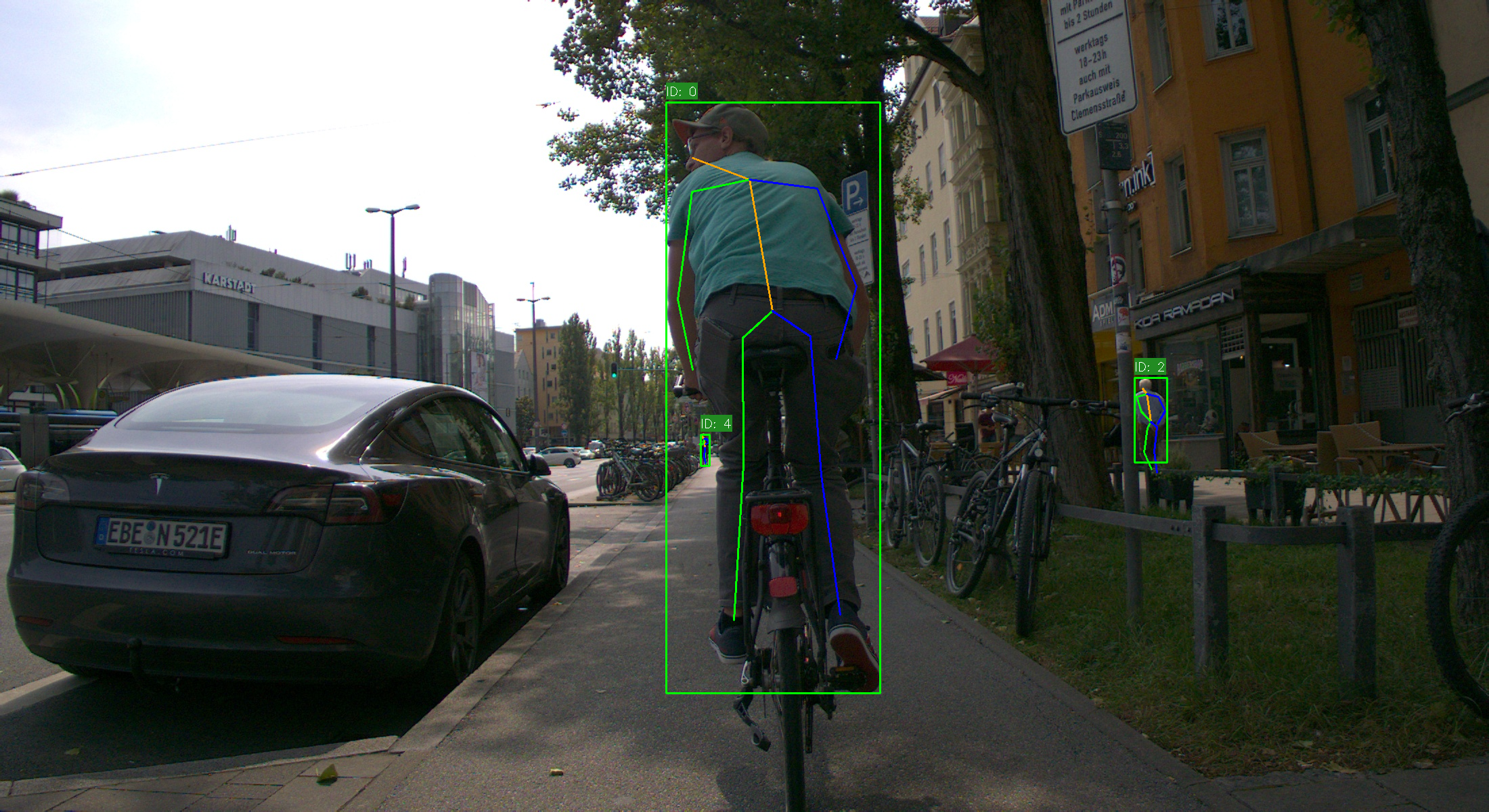} 
        \label{fig:glimpse-rgb3}
    \end{subfigure}



    \begin{subfigure}{0.32\linewidth}
        \includegraphics[width=\linewidth]{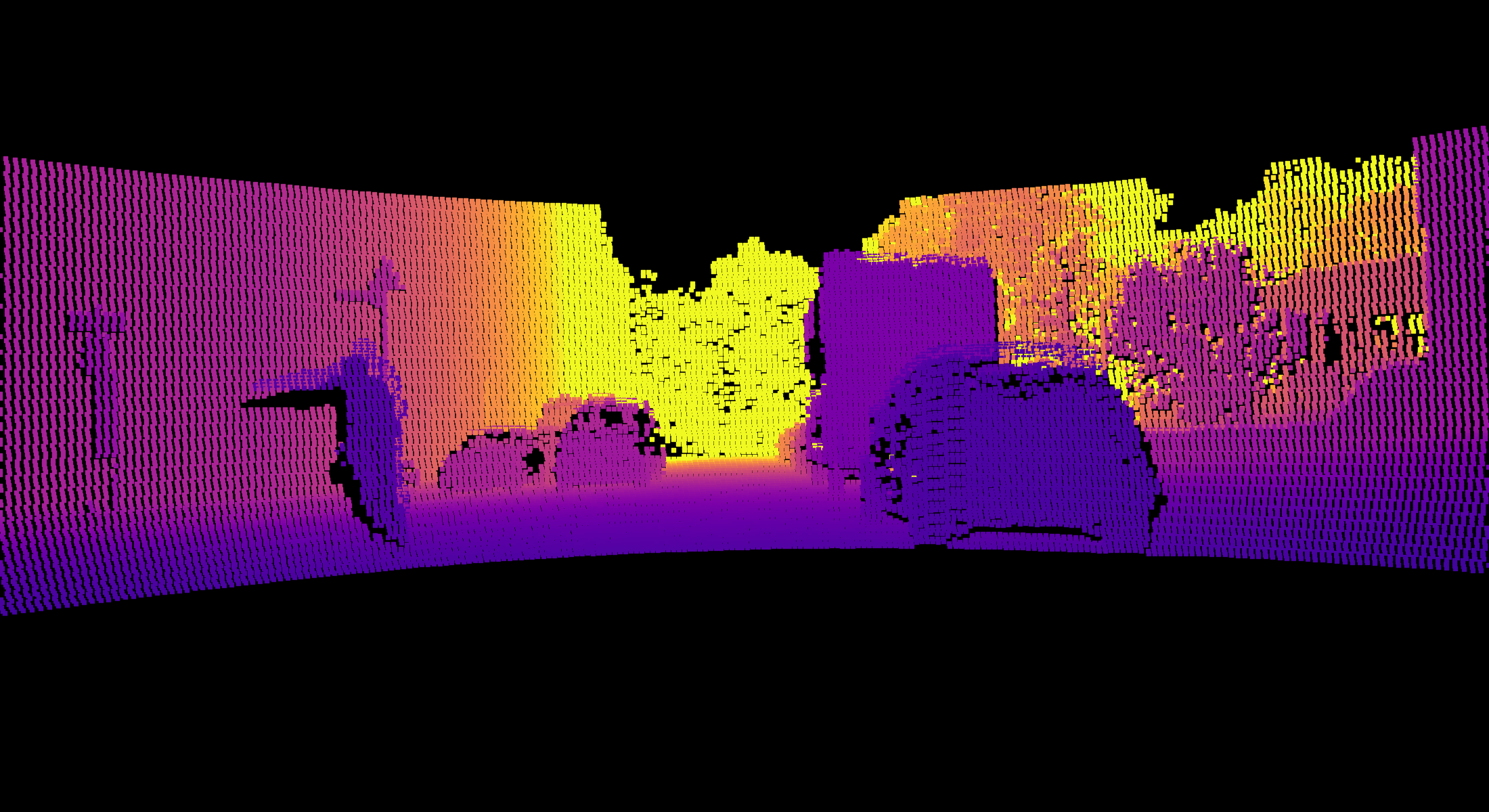} 
        \label{fig:glimpse-depth1}
    \end{subfigure}
    \hfill
    \begin{subfigure}{0.32\linewidth}
        \includegraphics[width=\linewidth]{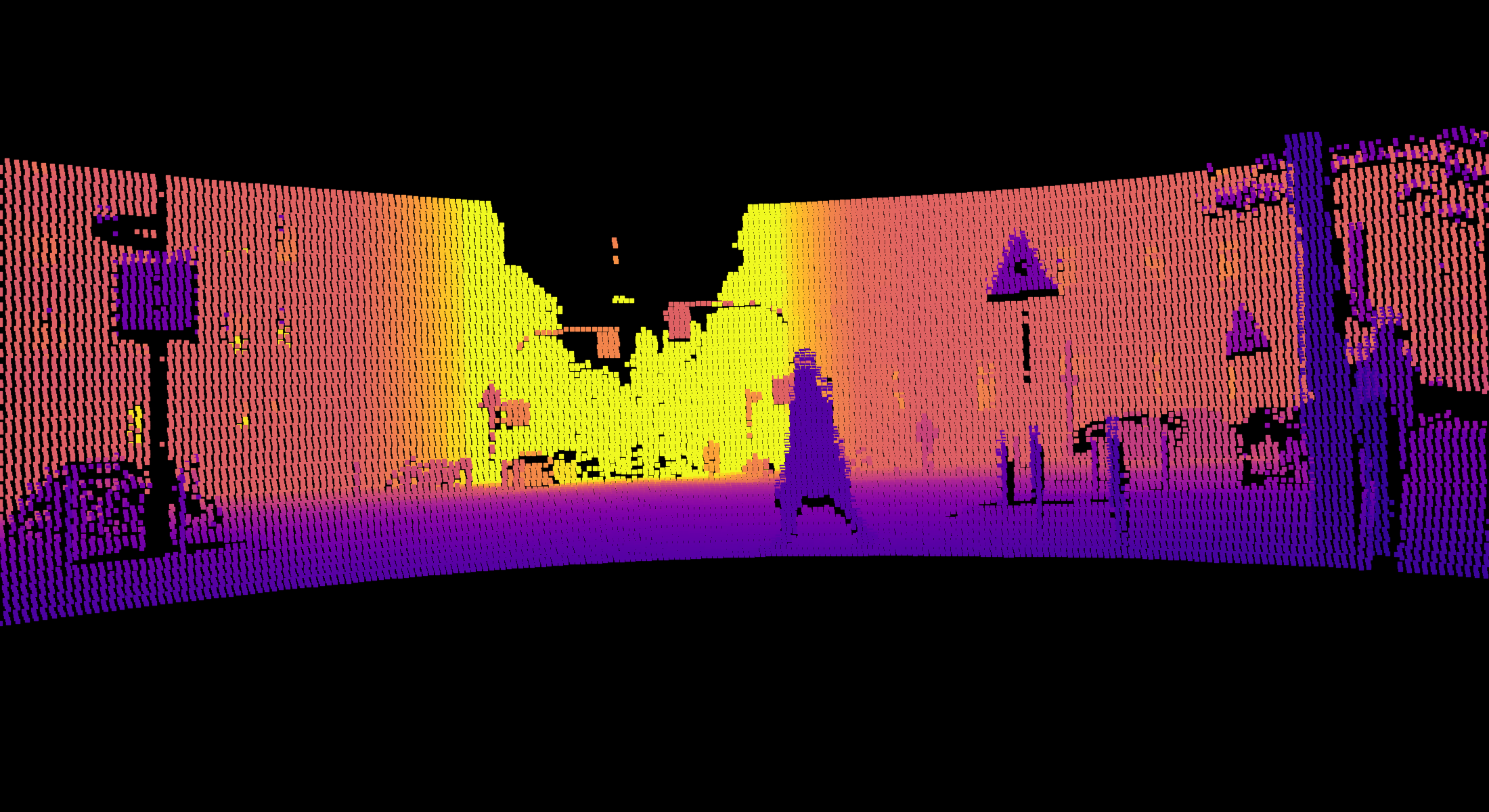} 
        \label{fig:glimpse-depth2}
    \end{subfigure}
    \hfill
    \begin{subfigure}{0.32\linewidth}
        \includegraphics[width=\linewidth]{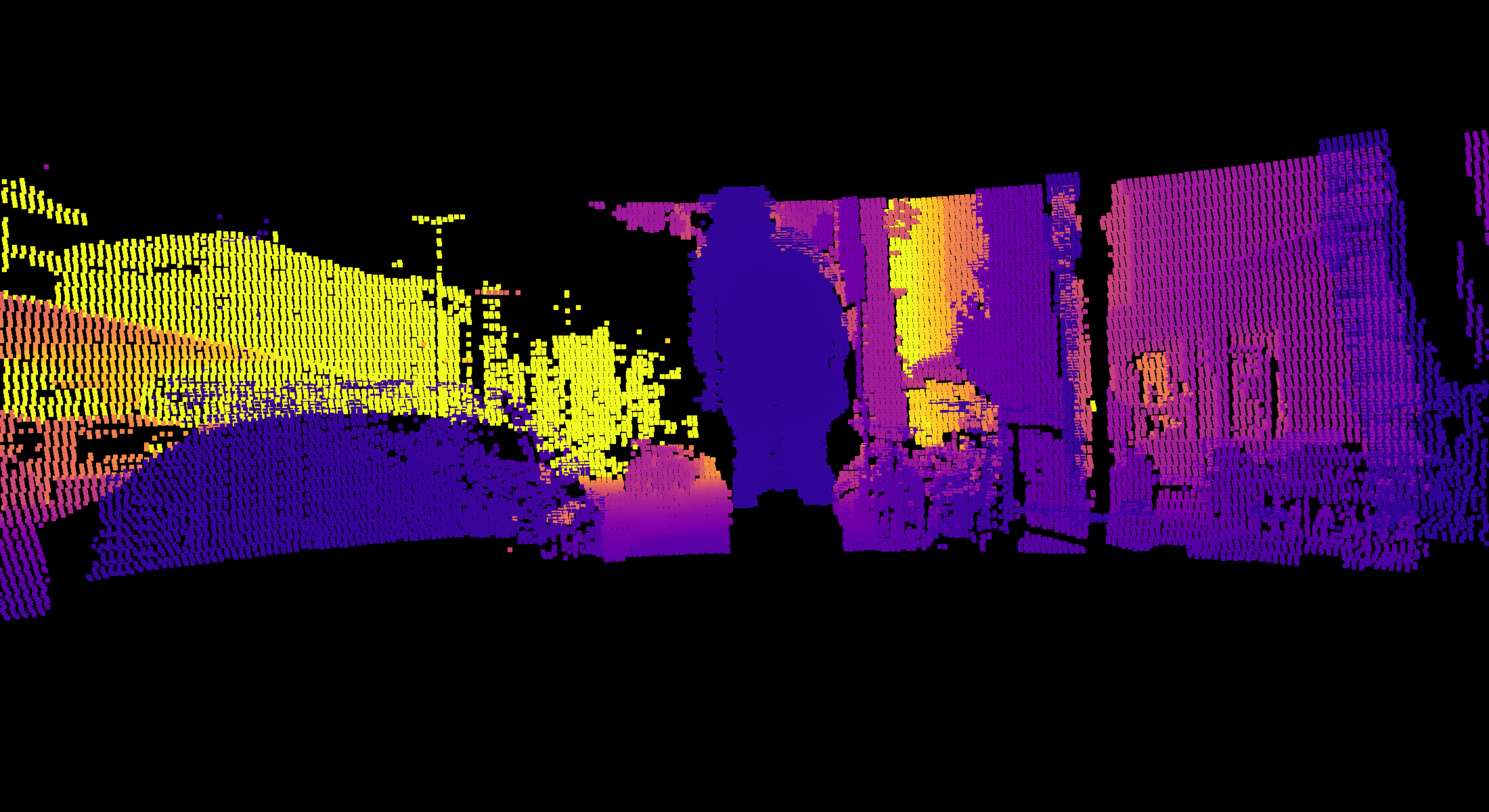} 
        \label{fig:glimpse-depth3}
    \end{subfigure}

    \caption{Qualitative examples from the \textit{BikeActions} dataset recorded with the FUSE-bike platform. The top row shows RGB camera views with projected 2D skeleton overlays for three distinct urban scenarios (hand sign in adverse lightning conditions, pedestrian crossing, narrow bicycle lane). The bottom row displays the corresponding sparse depth images from the long-range LiDAR, colorized to represent depth on a 0-50m scale.}
    \label{fig:glimpse-combined}
\end{figure*}

To address these challenges, we introduce a comprehensive framework bridging data acquisition, processing, and benchmarking. Implementing our novel \textit{FUSE-Bike} platform directly in the field, we curate \textit{BikeActions} (see Fig.~\ref{fig:glimpse-combined}), filling a critical data gap regarding interactions in dense VRU shared spaces. To our knowledge, this constitutes the first large-scale 3D human pose dataset of its kind, offering vital insights for autonomous vehicles and mobile robotics operating in complex urban environments.

Our contributions are threefold. 
First, we introduce \textit{FUSE-Bike}, the first fully open bicycle-mounted perception platform of its kind. Featuring a high-fidelity sensor suite with robust extrinsic calibration and hardware-level synchronization, it serves as a blueprint for the community to lower the barrier for ego-centric micro-mobility research in shared spaces.
Second, we present \textit{BikeActions}, a first dataset with \samplecount{} annotations captured from a cyclist's perspective; this unique viewpoint is critical for fine-grained VRU behavior modeling, directly serving to improve the safety and performance of AD systems and robots moving in shared spaces.
Third, we establish the first benchmark for cyclist-centric action recognition by evaluating multiple state-of-the-art skeleton-based models. To ensure reproducibility and fair comparison, we release our standardized data splits and the complete benchmark evaluation code, allowing future work to be directly evaluated against our results.

\section{Related Work}
\label{sec:related_work}

\subsection{Urban Data Collection Platforms}

The advancement of autonomous systems has been driven by large-scale data collection platforms, typically instrumented vehicles, designed to capture rich sensor data from real-world environments. Pioneering efforts like the KITTI dataset established the value of combining cameras and LiDAR for 3D perception tasks~\cite{geigerAreWeReady2012}. This approach was later scaled up by industry and academia, leading to comprehensive platforms that produced seminal datasets such as Waymo~\cite{sunScalabilityPerceptionAutonomous2020} and nuScenes~\cite{caesarNuScenesMultimodalDataset2020a}, which have become standard benchmarks for autonomous vehicle research.

While these platforms are technologically sophisticated, sensor platforms in urban environments are almost exclusively car-centric~\cite{burnettZeusSystemDescription2021},\cite{haselbergerJUPITERROSBased2022},\cite{karleEDGARAutonomousDriving2024}. 
This vehicle-based approach imposes a specific, high-elevation sensor viewpoint and inherently constrains data acquisition to roadways accessible by cars, often failing to capture the close-quarters, nuanced interactions involving VRUs.
To capture the world from a non-vehicular viewpoint, a smaller body of work has explored platforms ranging from wearable rigs to sensor-equipped bicycles, as detailed in Table~\ref{tab:sensor_platform_comparison}. Many of these efforts are limited to a single modality, such as the LiDAR-only SaBi3D~\cite{odenwaldSaBi3dALiDARPoint2024} or the camera-only CASR~\cite{fangIntentionRecognitionPedestrians2020}. Even fully multi-modal platforms often face significant technical hurdles; the helmet-based SLOPER4D~\cite{daiSLOPER4DSceneAwareDataset2023} lacks both sensor calibration and hardware time synchronization, while the Oxford RobotCycle backpack~\cite{robotcycle} only achieves partial Precision Time Protocol (PTP) synchronization across its sensors. Other projects, such as the AuRa cargo bike~\cite{sassAuRaDatasetVision2025} and BikeScenes-lidarseg~\cite{bikescenes-goren}, have highlighted the value of the VRU perspective but have so far only released datasets limited to a single sensor modality or task, such as semantic segmentation.

The field lacks a platform that combines the agility to navigate shared spaces with the rigorous sensor synchronization and calibration of an autonomous vehicle. We bridge this gap with \textit{FUSE-Bike}, the first fully open-source platform to bring automotive-grade perception standards to a cyclist's vantage point. This hardware foundation enables a first data contribution, \textit{BikeActions}. Comprising \samplecount{} annotated samples derived from 46,180 synchronized frames per sensor (one camera and two LiDARs), it stands as the largest multi-modal dataset of its kind, significantly surpassing comparable non-vehicular benchmarks in scale and fidelity. By openly releasing this high-resolution data capturing the dense, unregulated interactions unique to bike lanes and sidewalks, we provide the community with the first robust resource to model fine-grained VRU action classes with a precision previously unattainable.

\begin{table*}[t]
\caption{Comparison of non-vehicular sensor platforms and their target domains to FUSE-Bike (ours). \#f: number of frames. C/L/G/R: Camera/LiDAR/GNSS/Radar. Sync.: Software (SW) and Hardware (HW) Synchronization. Calib.: Extrinsic Calibration.}
\label{tab:sensor_platform_comparison}
\centering
\footnotesize
\begin{tabularx}{\textwidth}{@{}X r cccc l c l X @{}} 
\toprule
\textbf{Platform} & \textbf{\#f} & \multicolumn{4}{c}{\textbf{Sensors}} & \textbf{Platform} & \textbf{Sensor} & \textbf{Time} & \textbf{Domain} \\
\cmidrule(lr){3-6} 
& & \textbf{C} & \textbf{L} & \textbf{G} & \textbf{R} & & \textbf{Calib.} & \textbf{Sync.} & \\
\midrule
SaBi3D~\cite{odenwaldSaBi3dALiDARPoint2024} & 4.8k & \xmark & \cmark & \xmark & \xmark & Bike & --- & --- & Urban SLAM \\
RadBike~\cite{renLiDARaidInertialPoser2023} & --- & \cmark & \xmark & \xmark & \cmark & Bike & \cmark & SW & Odometry \\
RobotCycle~\cite{robotcycle} & --- & \cmark & \cmark & \cmark & \xmark & Backpack & \cmark & HW & Urban Mapping \\
SLOPER4D~\cite{daiSLOPER4DSceneAwareDataset2023} & 32k & \cmark & \cmark & \xmark & \xmark & Helmet & \xmark & SW & Motion Capture \\
CASR~\cite{fangIntentionRecognitionPedestrians2020} & 40k & \cmark & \xmark & \xmark & \xmark & Static & --- & --- & Ped. Intent \\
AuRa~\cite{sassAuRaDatasetVision2025} & 0,5k & \cmark & \cmark & \cmark & \cmark & Cargo Bike & \cmark & HW & AD / Detection \\
BikeScenes~\cite{bikescenes-goren} & 3k & \cmark & \cmark & \cmark & \xmark & Bike & \cmark & SW & Lid. Segm. \\
\midrule
\textbf{FUSE-Bike} & \textbf{46k} & \textbf{\cmark} & \textbf{\cmark} & \textbf{\cmark} & \textbf{\xmark} & \textbf{Bike} & \textbf{\cmark} & \textbf{HW} & \textbf{AD / Act. Rec.} \\
\bottomrule
\end{tabularx}
\end{table*}

\subsection{Action Recognition and Driving Datasets}

The task of human action recognition began with methods focused on classifying actions from static images~\cite{stanford40,pascalvoc}, relying on pose and context cues~\cite{image_action_ref}. However, to capture the essential temporal dynamics of motion, the field quickly transitioned towards video-based approaches. Seminal large-scale datasets of trimmed video clips, such as Kinetics~\cite{kinetics}, were instrumental in this shift, enabling the development of deep learning models that established paradigms for 2D and 3D spatiotemporal feature learning. In parallel, skeleton-based action recognition emerged as an efficient alternative leveraging 2D or 3D human pose data as a compact and robust representation, focusing purely on human motion while being invariant to distracting factors like background and appearance.

Popular datasets such as NTU RGB+D~\cite{ntu-rgbd}, NTU RGB+D 120~\cite{ntu-rgbd120}, PKU-MMD~\cite{pkummd}, and NW-UCLA~\cite{ucla} provide skeleton along with RGB, depth, and infrared for diverse human action recognition for daily tasks and multi-subject interaction. These datasets are recorded indoors, with staged actors, and usually with static RGB-D cameras. Kinetics uses RGB videos from Youtube with OpenPose generated skeleton. 
In comparison to these generic human action recognition datasets, JAAD~\cite{jaad}, STIP~\cite{stip}, PIE~\cite{pie} capture and annotate road user action with the goal of pedestrian intent prediction, but limit their scope primarily to binary crossing scenarios. More recently, datasets like ROAD-Waymo~\cite{road-waymo-github} have provided action labels on top of a large-scale automotive dataset.

However, a critical gap remains. Even the most relevant existing datasets are capture from a vehicles viewpoint, subsequently missing to capture the unique, close-quarters perspective of a cyclist or pedestrian, which is essential for modeling subtle interaction in shared urban spaces. Hence, we introduce a multi-modal dataset with annotated VRU (pedestrian and cyclist) skeleton and actions relevant for AD platforms for predicting and planning safe maneuvers.

\subsection{Skeleton based Action Recognition}
Skeleton based action recognition classify human action based on joint coordinates and connectivity in the skeleton. Skeleton action recognition approaches can be divided into four categories: CNN-based, RNN-based, GCN-based, and Transformer-based methods. Due to the sequential and continuous nature of the skeleton action samples, early works utilizes Recurrent Neural Networks (RNNs) along with LSTM and GRU units to be able to handle longer action samples. Due to advancements of Convolutional Neural Networks (CNNs), some works adopted 2d and 3D CNN-based methods for action classification. Recent approaches also used skeleton data composed of joints and bones, which can be directly modeled into a graph with joints as vertices and bones as edges. As such, graph based networks such as  Graph Neural Networks (GNNs) and Graph Convolution Networks (GCNs) attained dominated this field. However, these models fail to capture relation between physically distant joints as they are limited to local spatial-temporal neighborhoods~\cite{skateformer}, \cite{action-survey}. Recent approaches focus on transformer-based approaches, which are more capable in capturing global topology and relation between physically distant joints. Current state-of-the-art models integrate GCN with transformer variants with reduced computation and memory cost. More detailed analysis of these approaches can be found in~\cite{action-survey}.






\section{FUSE-Bike Perception Platform}
\label{sec:fuse_bike_platform}
\begin{figure}[htbp]
    \centering
    \includegraphics[width=0.9\linewidth]{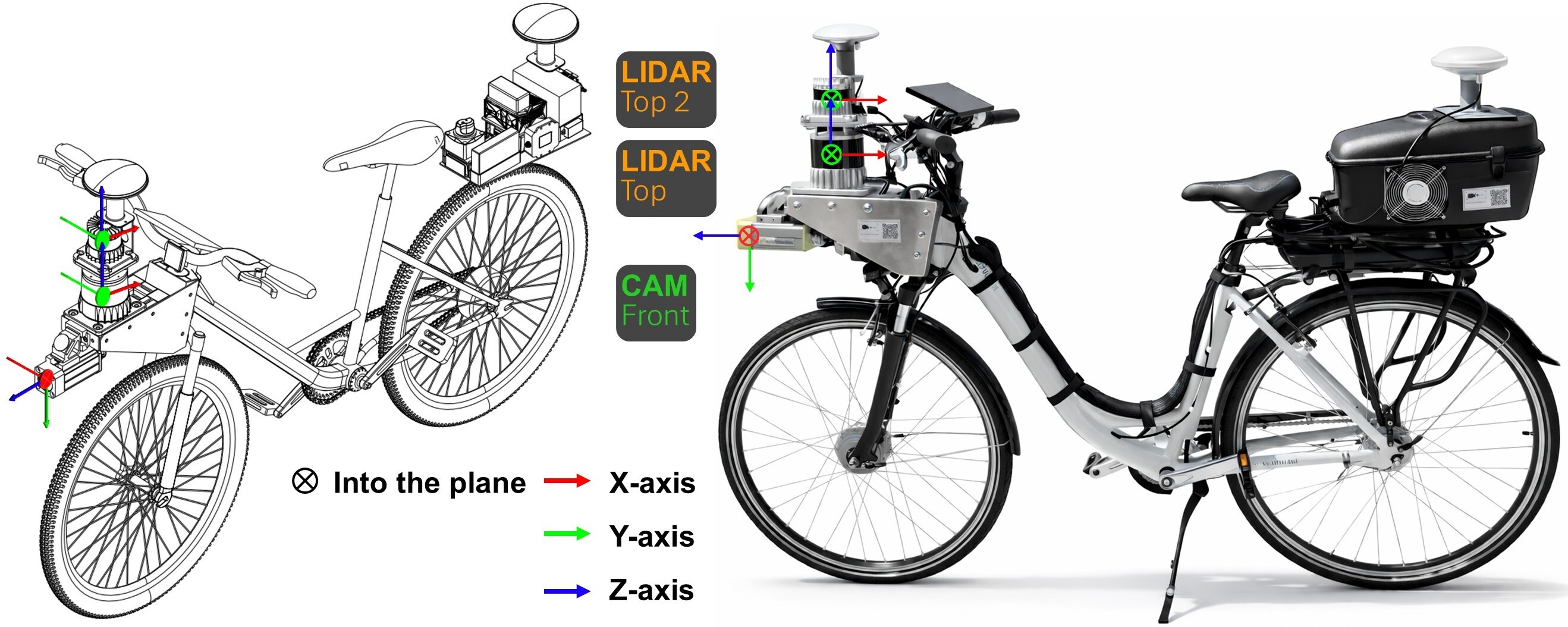}
    \caption{The FUSE-Bike hardware prototype and its corresponding CAD model. The design features a rigid front sensor mount with stacked LiDARs and a camera, with the main electronics housed in a rear-mounted case.}
    \label{fig:bike_picture}
\end{figure}

To address the limitations of existing data collection systems, we designed and built the FUSE-Bike, an open and accessible multi-modal perception platform. The core design principle is providing a blueprint for a reproducible and VRU-centric data collection system, allowing the extension of it by the research community. All aspects of the platform, from hardware design to software integration, will be released publicly. The platform's CAD model and hardware are depicted in Fig.~\ref{fig:bike_picture}. 

\subsection{Hardware Configuration}
FUSE-Bike is built on a rugged, electrically assisted bicycle frame chosen to support the system's weight and power requirements. All core electronics are housed in a rear-mounted, actively-cooled case for compactness and protection, including the 360Wh battery pack, dedicated DC/DC converters, and 6TB of data storage. An NVIDIA Jetson AGX Orin serves as the central compute unit, connected via a 2.5Gbit/s Ethernet switch to the front-mounted sensor suite. This suite provides a comprehensive environmental view, featuring a vertically stacked LiDAR tower (Ouster OS2-128 for long-range, OS0-128 for near-field), a high-resolution Basler monocular camera, and a dual-antenna Septentrio RTK-GNSS module for robust, high-precision positioning and heading estimation. All components are rigidly mounted using custom CNC-machined and 3D-printed brackets to ensure mechanical stability. Finally, an 8-inch touchscreen on the handlebar provides a human-machine interface (HMI) for mobile system monitoring and control.
The final prototype in Fig.~\ref{fig:bike_picture} emerged from several iterative design and integration cycles, each aimed at optimizing cross-component compatibility, structural integrity, and real-time performance under dynamic urban cycling conditions. The sensor components are further detailed in Table~\ref{tab:fuse_bike_sensors}.

\begin{table}[htbp]
    \caption{Sensor specifications for the FUSE-Bike platform.}
    \label{tab:fuse_bike_sensors}
    \centering
    \begin{tabular}{|l|m{9cm}|}
        \hline
        \textbf{Sensor} & \textbf{Details} \\
        \hline
        \hline
        1x Camera & Basler Ace2 pro GigE, 12Bit RGGB, 60Hz max. with 10Hz capture freq., 1/1.8'' CMOS sensor, 2200 $\times$ 1200 resolution \\
        \hline
        1x OS2 & Ouster OS2-128, 128 beams, 10 Hz capture freq., 360$^{\circ}$ horiz. FOV, $\pm{}11.25^{\circ}$ vert. FOV, $200$m @ 10\% range \\
        \hline
        1x OS0 & Ouster OS0-128, 128 beams, 10Hz capture freq., 360$^{\circ}$ horiz. FOV, $\pm{}45^{\circ}$ vert. FOV, $35$m @ 10\% \\
        \hline
        1x GNSS & Septentrio AsteRx-m3 Pro+, dual-antenna, GPS, IMU, AHRS, $0.1^{\circ}$ heading accuracy, $0.05^{\circ}$ roll/pitch accuracy, 10mm RTK positioning accuracy, 100Hz update rate \\
        \hline
    \end{tabular}
\end{table}


\subsection{Sensor Calibration}

Spatial alignment across all sensors is achieved through a multi-stage calibration process, resulting in a tree of static transforms managed by ROS2 TF2 with the long-range Ouster OS2 LiDAR as the bicycle's origin frame (\texttt{base\_link}). The camera's intrinsic parameters are first determined using a checkerboard pattern, yielding the camera matrix $\mathbf{K}$ (Eq.~\eqref{eq:intrinsics}):
\begin{align}
    \mathbf{K} &= \begin{bmatrix} f_x & s & c_x \\ 0 & f_y & c_y \\ 0 & 0 & 1 \end{bmatrix} \label{eq:intrinsics}
\end{align}    

Next, the extrinsic transformations between sensor frames are found. The static transform from the OS2 origin to the camera ($\mathbf{T}_{\text{cam} \leftarrow \text{os2}}$) is estimated using LiDARTag markers~\cite{lidartag}, 
while the transform from the OS2 to the near-field OS0 LiDAR ($\mathbf{T}_{\text{os0} \leftarrow \text{os2}}$) is found using a mapping-based approach~\cite{tier4-calibration-tools}, comparing planar and vertical structures. This calibrated TF2 tree allows for the projection of any 3D point into the image plane. For example, a point from the world frame, $\mathbf{P}_w$, is projected to a pixel in homogeneous coordinates $\mathbf{p}$ with depth $\mathbf{\lambda}$ using its full transform chain (Eq.~\eqref{eq:full_projection_chain}):
\begin{align}
    \lambda \mathbf{p} &= \mathbf{K} \, [\mathbf{I}|\mathbf{0}] \, \mathbf{T}_{\text{cam} \leftarrow \text{os2}} \, \mathbf{T}_{\text{os2} \leftarrow \text{world}} \, \mathbf{P}_w \label{eq:full_projection_chain}
\end{align}
Similarly, a point from the near-field LiDAR frame, $\mathbf{P}_{\text{os0}}$, is projected using its respective chain (Eq.~\eqref{eq:os0_projection}): 
\begin{align}
    \lambda \mathbf{p} &= \mathbf{K} \, [\mathbf{I}|\mathbf{0}] \, \mathbf{T}_{\text{cam} \leftarrow \text{os2}} \, (\mathbf{T}_{\text{os0} \leftarrow \text{os2}})^{-1} \, \mathbf{P}_{\text{os0}} \label{eq:os0_projection}
\end{align}
This entire set of transforms is then globally refined through SLAM-aided adjustments to minimize reprojection error.

\subsection{Time Synchronization}
\begin{figure}[h!]
    \centering
    \includegraphics[width=0.6\linewidth]{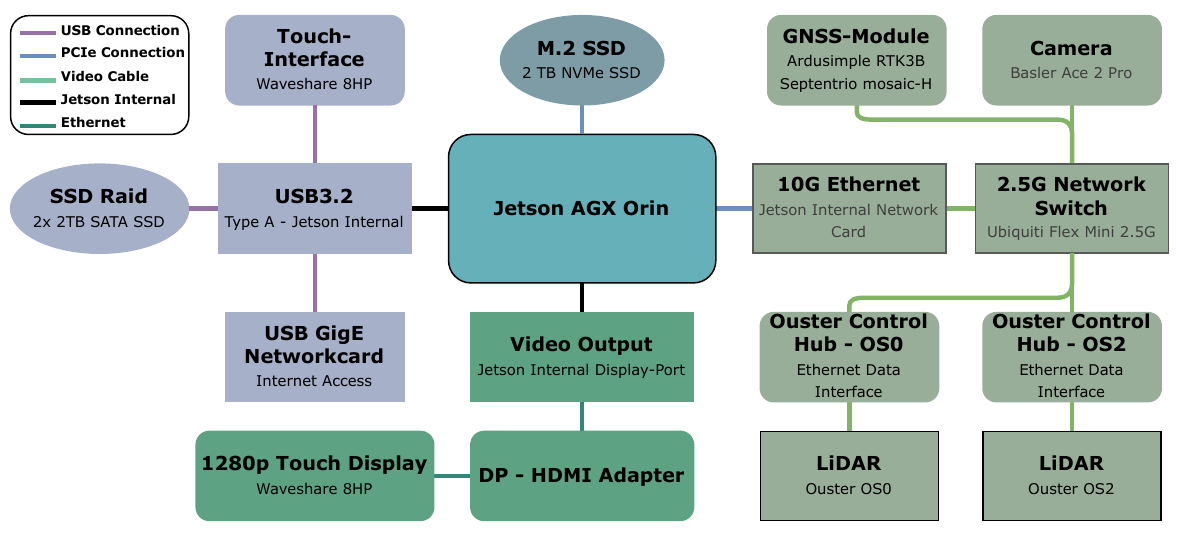}
    \caption{System architecture of the FUSE-Bike, showing the data flow between the Jetson compute unit, the PTP-synchronized sensors, and other key electronics.}
    \label{fig:system_architecture}
\end{figure}

To ensure temporal consistency, all data streams are synchronized at the hardware level using the Precision Time Protocol (PTP). The Septentrio GNSS receiver operates as the PTP master, distributing a unified, high-precision clock signal to the LiDARs, camera, and Jetson over the Ethernet network. This guarantees that all sensor data shares a common time base with microsecond-level accuracy, which is critical for reliable multi-modal fusion. The wiring between compute unit and hardware is shown in Fig.~\ref{fig:system_architecture}.

\section{BikeActions Dataset}
\label{sec:bikeactions_dataset}

\subsection{Data Acquisition \& Processing}
\label{subsec:data_acquisition}

Our data collection was conducted in various urban and residential districts of Munich, Germany, using the FUSE-Bike platform detailed in Section~\ref{sec:fuse_bike_platform}. Recordings were captured exclusively during daytime hours under a variety of lighting conditions, ranging from bright sunlight to overcast skies, and in clear, dry weather. To ensure a rich variety of VRU behaviors, we not only recorded common traffic behavior but also included scenarios critical for autonomous systems, including intersections, designated bike lanes, shared road spaces, and pedestrian crossings. The data was gathered through a combination of naturalistic cycling, where the rider navigated traffic normally, and targeted following, where interesting VRU subjects were observed from a safe distance to capture longer, continuous action samples.

The continuous, long-duration recordings were then processed and curated. First, they were split into non-overlapping, manageable chunks, or "parts", each containing roughly 200 frames (20 seconds of data). These parts were then passed through an automated pre-computation pipeline to generate the necessary inputs for annotation. For each frame, this pipeline extracted the raw camera images and LiDAR-derived depth maps. Subsequently, we ran a state-of-the-art detector and tracker to generate unique person IDs and 2D bounding boxes, as well as a 3D human pose estimator to generate initial 3D keypoint estimations. Finally, we curated these parts for annotation and prioritized sequences containing clearly visible VRUs with high-quality tracking for a significant duration.

To support efficient annotation, we pre-rendered visualization videos for each part, including colorized depth maps and images overlaid with bounding boxes and 2D skeletons, yielding a processed package of raw data, pre-computed poses, and visualizations that serve as direct input to our semi-automated action annotation tool.
  
\subsection{Data curation}
Existing annotation tools lack support for simultaneous multi-subject labeling, frame-level temporal precision, and custom hierarchical class taxonomies, requirements essential for our urban VRU action dataset. To address these limitations, we developed a custom annotation tool that combines automatic temporal boundary suggestions with interactive manual refinement. The tool streamlines the labeling workflow by automatically populating initial start and end frames based on complete skeleton tracking, allowing annotators to refine these boundaries and verify action continuity via frame-by-frame scrubbing before assigning class labels. 


To ensure consistency across all annotations, we established a strict labeling protocol for action samples. Each action sample is defined by a single action label applied to a continuous sequence of frames where a VRU performs that same action. The action labels are "Walking" for a walking pedestrian, "Standing" for a standing pedestrian, "Cycling" for a moving cyclist, "Cycling: Left" for a cyclist indicating a left hand signal, and "Cycling: Right" for a cyclist indicating a right hand signal. Due to under-representation, classes such as running pedestrians or waiting cyclists have been removed. If a VRU changes their action, the sequence is split into distinct samples; e.g. an uninterrupted sequence of `cycling` $\rightarrow$ `left hand signal` $\rightarrow$ `cycling` results in three separate, consecutive annotations. For a hand signal to be classified as such, the corresponding arm must be visibly raised. Furthermore, to ensure that each sample contains sufficient temporal information for model training, we enforce a minimum duration, requiring every annotated action sample to consist of at least 10 consecutive frames of the fully visible tracked VRU.

    
\subsection{Dataset Statistics}
\begin{figure}[h!]
    \centering
    \begin{subfigure}[t]{0.59\textwidth}
        \includegraphics[width=\linewidth]{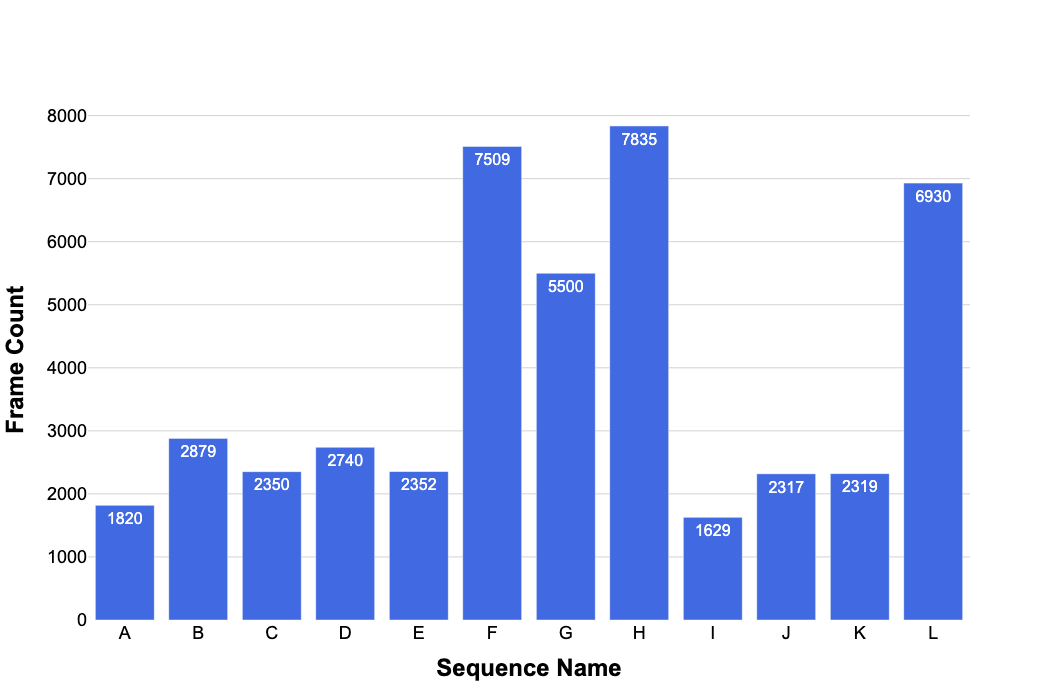}
        \caption{Total frames.}
        \label{fig:dataset_stats_frames}
    \end{subfigure}
    \hfill 
    \begin{subfigure}[t]{0.40\textwidth}
        \includegraphics[width=\linewidth]{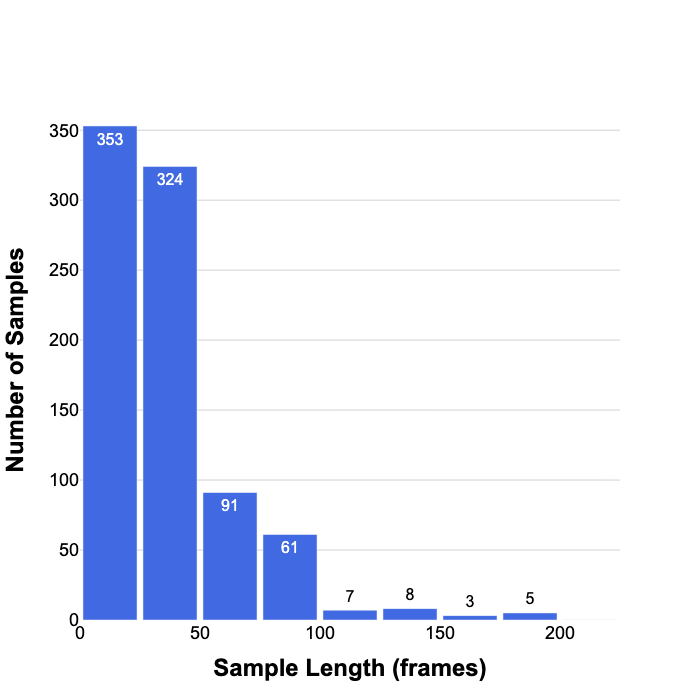}
        \caption{Sample length.}
        \label{fig:dataset_stats_length}
    \end{subfigure}
    
    \caption{Statistics of the \textit{BikeActions} dataset. 
    (\subref{fig:dataset_stats_frames}) total number of raw frames per sequence;
    and (\subref{fig:dataset_stats_length}) Histogram showing the distribution of action sample duration.
    }
    \label{fig:tool_and_stats}
\end{figure}

The final \textit{BikeActions} dataset spans 12 unique sequences with a cumulative duration of 1.3 hours. It yields 46,180 synchronized frames per sensor. The distribution of these frames across each recording session (A-L) is visualized in Fig.~\ref{fig:dataset_stats_frames}. From this raw data, our annotation process yielded a total of \samplecount{} high-quality annotated action samples across five classes, evenly split into 70-15-15 train-validation-test. Furthermore, a histogram of all sample lengths is presented in Fig.~\ref{fig:dataset_stats_length}, revealing a focus on short, atomic actions with an overall average duration of $36.2$ frames.

\begin{table*}[htbp]
    \caption{Summary of the five annotated action classes in \textit{BikeActions}.}
    \label{tab:class_summary}
    \centering
    \begin{tabular}{c|l|l|r|r}
        \toprule
        \textbf{Class ID} & \textbf{Class Label} & \textbf{Short Label}& \textbf{Samples} & \textbf{Avg. Length} \\
        \midrule
        1 & Walking         & walk  & 330   & 26.5 \\
        2 & Standing        & stand & 122   & 27.0 \\
        3 & Cycling         & bike  & 271   & 54.8 \\
        4 & Cycling: Left   & left  & 62    & 31.1 \\
        5 & Cycling: Right  & right & 67    & 30.4 \\
        \bottomrule
    \end{tabular}
\end{table*}

\section{Experiments and Results}
\label{sec:experiments}
\subsection{Benchmark Models}

We implement and benchmark five state-of-the-art human action classification models on our dataset across two skeleton modalities: joints and bones. As bones encode the physical connectivity among joints of human body, models usually perform better with bone modality than joint~\cite{skateformer,hd-gcn}. Therefore, 10 different models are evaluated in the benchmark. Among the five different types of skeleton based human action classification models we choose GCNs and Transformer based models due to their higher performance over CNN and RNN based models.

\paragraph{GCN-based Action Recognition}
A skeleton sequence, corresponding to a single action is sampled to maintain a consistent temporal window T and represented as $\mathbf{X} \in \mathbb{R}^{C_{in} \times T \times V}$, where V is the number of joint nodes and $C_{in}$ is the dimensionality of data representing a single joint. GCN takes as input a feature map $\mathbf{F}_{in} \in \mathbb{R}^{C \times T \times V}$ and updates joint features over predefined graph subsets and outputs $\mathbf{F}_{out} \in \mathbb{R}^{C' \times T \times V}$, as in 
\begin{equation}
    \mathbf{F}_{out} = \sum_{s \in S} \mathbf{\hat{A}}_s \mathbf{F}_{in} \Theta_s,
\end{equation}
where, $S$ is graph subsets, $\Theta_s$ is pointwise convolution and $\mathbf{\hat{A}}$ is normalized adjacency matrix. GCN extracted features are finally passed to a classification head after global average pooling over joints and time. We pick HD-GCN~\cite{hd-gcn}, CTR-GCN~\cite{ctr-gcn}, and Neural Koopman Pooling model~\cite{koopman} for our benchmark.

\paragraph{Transformer-based Action Recognition}
Transformer based action recognition models represent skeleton action sequence in a similar manner. The input skeleton data $\mathbf{X} \in \mathbb{R}^{T \times V \times C_{in}}$ is projected to higher-dimension feature space using linear layers before adding positional embedding. The output features from the transformer blocks are then passed to classification head for action recognition. We choose Hyperformer\cite{hyperformer} and Skateformer~\cite{skateformer} models for setting baselines on our dataset.

\subsection{Implementation Details}
All experiments were performed on a single NVIDIA RTX 4090 GPU. For benchmarking, we adapted the models to be compatible with our dataset with minimal changes to the core methodology. To ensure fair comparison across models, we use all 20 vertices and crop or pad the sequence to 64. We follow original implementation for choosing other model hyperparameters. Additionally, classifying the five distinct pedestrian and cyclist actions provides a motion-based consistency check that can further confirm or refine VRU detections.



\subsection{Benchmark Analysis}

\begin{figure*}[h]
\centering

\begin{subfigure}{\textwidth}
    \centering
    \begin{minipage}[c]{0.78\linewidth}
        \includegraphics[width=\linewidth]{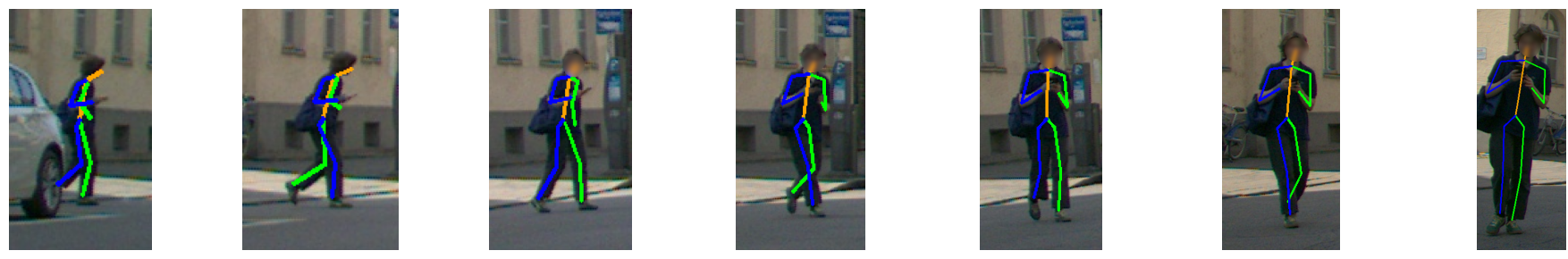}
    \end{minipage}
    \hfill
    \begin{minipage}[c]{0.18\linewidth}
        \centering
        \textbf{Class: Walk}
    \end{minipage}
\end{subfigure}

\vspace{0.5em}

\begin{subfigure}{\textwidth}
    \centering
    \begin{minipage}[c]{0.78\linewidth}
        \includegraphics[width=\linewidth]{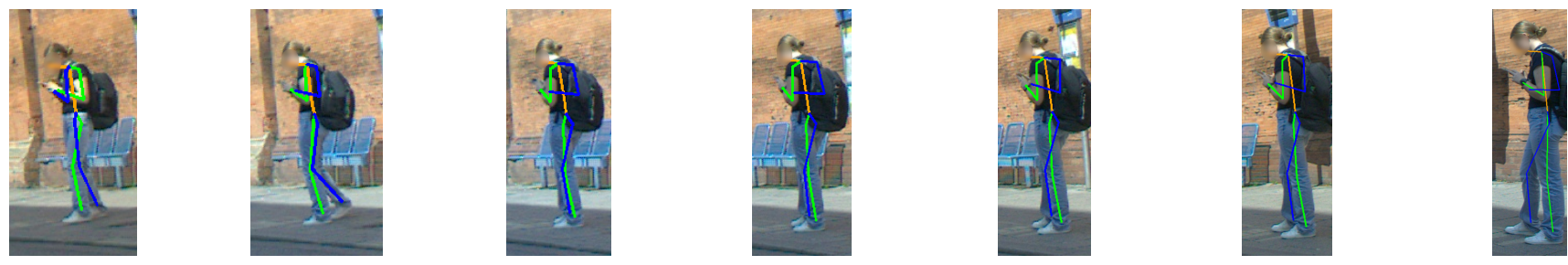}
    \end{minipage}
    \hfill
    \begin{minipage}[c]{0.18\linewidth}
        \centering
        \textbf{Class: Stand}
    \end{minipage}
\end{subfigure}

\vspace{0.5em}

\begin{subfigure}{\textwidth}
    \centering
    \begin{minipage}[c]{0.78\linewidth}
        \includegraphics[width=\linewidth]{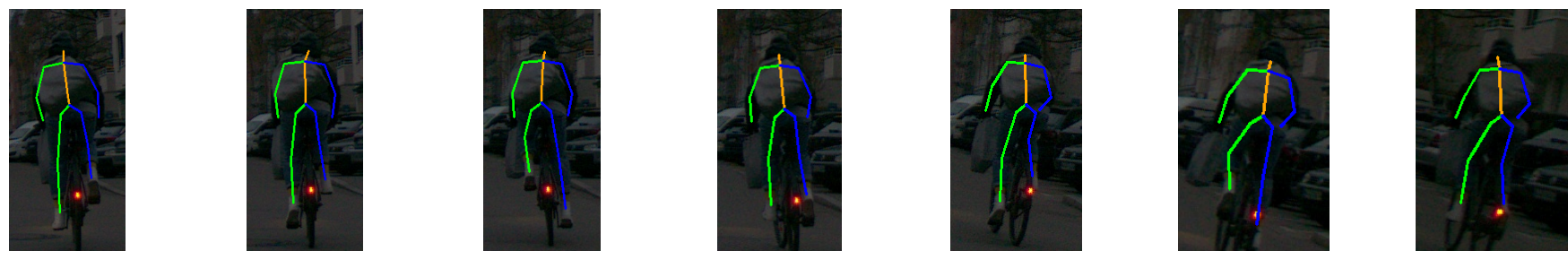}
    \end{minipage}
    \hfill
    \begin{minipage}[c]{0.18\linewidth}
        \centering
        \textbf{Class: Bike}
    \end{minipage}
\end{subfigure}

\vspace{0.5em}

\begin{subfigure}{\textwidth}
    \centering
    \begin{minipage}[c]{0.78\linewidth}
        \includegraphics[width=\linewidth]{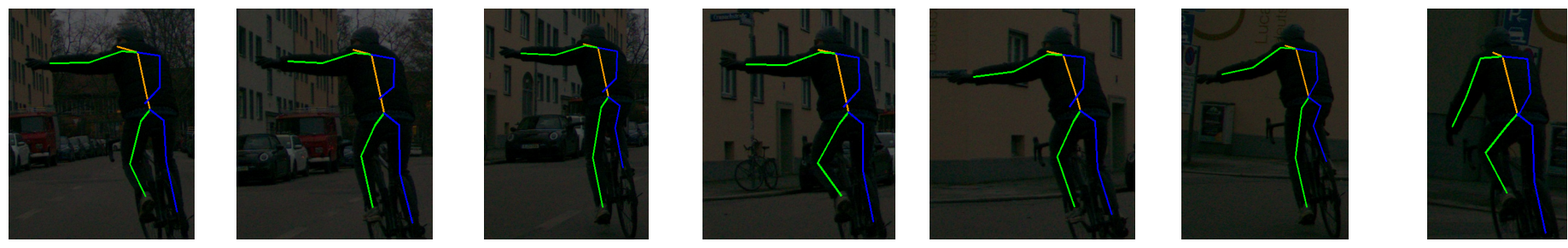}
    \end{minipage}
    \hfill
    \begin{minipage}[c]{0.18\linewidth}
        \centering
        \textbf{Class: Left}
    \end{minipage}
\end{subfigure}

\vspace{0.5em}

\begin{subfigure}{\textwidth}
    \centering
    \begin{minipage}[c]{0.78\linewidth}
        \includegraphics[width=\linewidth]{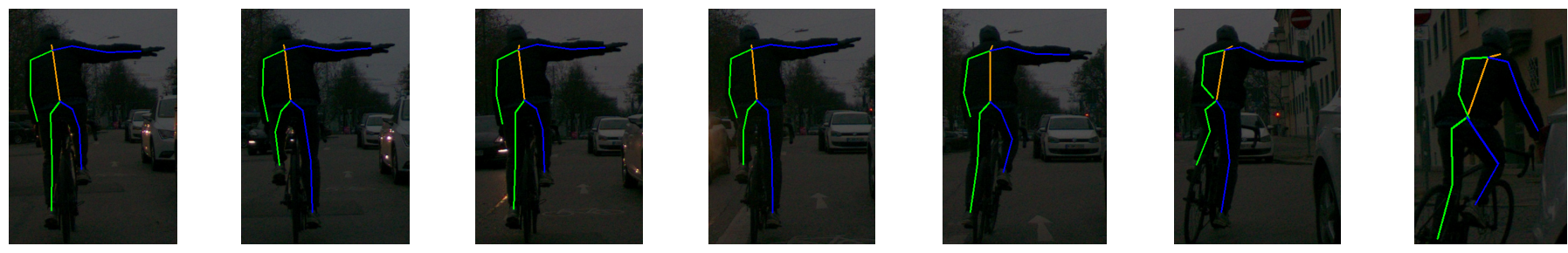}
    \end{minipage}
    \hfill
    \begin{minipage}[c]{0.18\linewidth}
        \centering
        \textbf{Class: Right}
    \end{minipage}
\end{subfigure}

\caption{Qualitative action recognition samples. Each row displays the sequence (left) alongside the corresponding class label (right). The skeleton is color-coded: right side in blue, left side in green, and central joints in orange.}
\label{fig:action_recognition_results}
\end{figure*}

\begin{figure}[h]
    \centering
    \begin{subfigure}[b]{0.42\linewidth}
        \includegraphics[width=\linewidth]{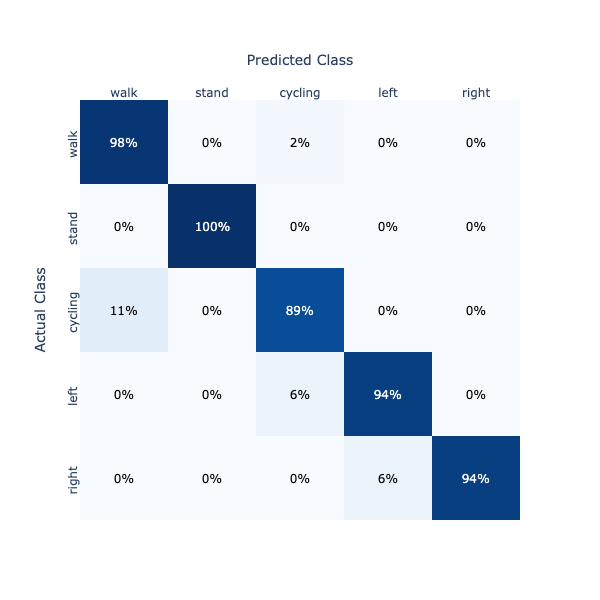}
        \caption{Hyperformer Bone}
    \end{subfigure}
    \hspace{0.02\linewidth} 
    \begin{subfigure}[b]{0.42\linewidth}
        \includegraphics[width=\linewidth]{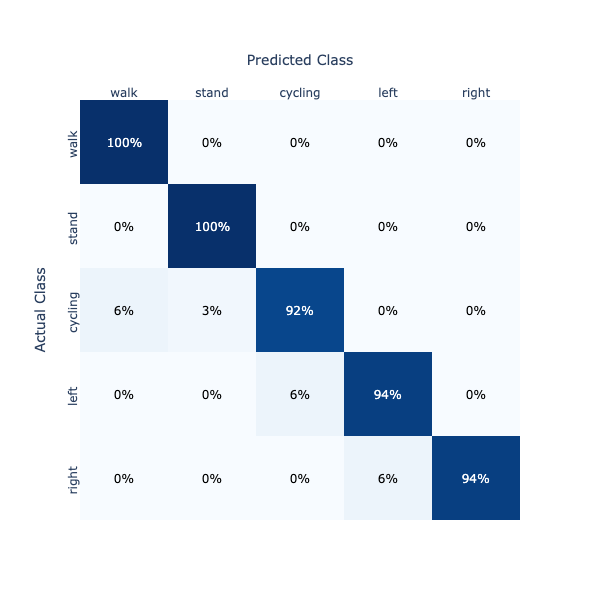}
        \caption{Hyperformer Joint}
    \end{subfigure}

    \caption{Confusion matrices for the best joint and bone models.}
    \label{fig:conf_matrix_all}
\end{figure}

We apply left-right sequence mirroring augmentation to account for symmetric human actions observed from the bicycle platform, effectively doubling the \textit{Cycling: Left} and \textit{Cycling: right} samples. Table~\ref{tab:comparison_aug} compares the accuracy (\%) of HD-GCN~\cite{hd-gcn}, CTR-GCN~\cite{ctr-gcn}, Koopman~\cite{koopman}, Hyperformer~\cite{hyperformer}, and SkateFormer~\cite{skateformer} on joint and bone modalities with this augmentation. Hyperformer achieves the highest accuracy on joint data (\textbf{96.15\%}) and bone data (\textbf{94.62\%}). The qualitative results from Hyperformer model trained on joint modality is shown in Fig.~\ref{fig:action_recognition_results}. The confusion matrices in Fig.~\ref{fig:conf_matrix_all} show that in most cases misclassifications stay in the superclass of cyclists or pedestrians respectively. These results indicate that Hyperformer effectively models joint-level and bone-level features. However, the differences among different approaches are very small considering the size of the \textit{val} split and may stem from hyperparameter choices. 

\begin{table*}[h]
\centering
\caption{Accuracy (\%) across different methods and the joint and bone modalities with left-right mirroring augmentation.}
\label{tab:comparison_aug}
\begin{tabular}{l|c|c|c|c|c}
\hline
 & HD-GCN~\cite{hd-gcn} & CTR-GCN~\cite{ctr-gcn} & Koopman~\cite{koopman} & Hyperformer~\cite{hyperformer} & SkateFormer~\cite{skateformer} \\
\hline
\hline
Joint & 66.92 & 93.08 & 92.31 & \textbf{96.15} & 95.38 \\
Bone  & 90.77 & 89.23  & 92.31 & \textbf{94.62} & 93.85 \\
\hline
\end{tabular}
\end{table*}

\section{Discussion and Outlook}

\paragraph{Discussion.}
Our benchmark results validate that \textit{BikeActions} is well-suited for training state-of-the-art skeleton-based models. Their strong performance suggests that 3D skeletons are a robust representation for VRU action classification, effectively disentangling motion from complex urban backgrounds. This is highly relevant for both autonomous vehicles and mobile robotics, as the FUSE-Bike's ground-level perspective captures close-range interactions that high-elevation automotive sensors miss. 
Furthermore, the annotation process highlighted the inherent "long-tail" problem of real-world data; the natural infrequency of critical actions like \textit{running} demands that future models learn effectively from imbalanced distributions.

\paragraph{Outlook.}
The FUSE-Bike platform and \textit{BikeActions} dataset enable several avenues for future research. A primary goal is to adapt the benchmarked models for real-time, onboard execution, transforming the FUSE-Bike into a reproducible benchmark for embodied AI and behavioral modeling for micromobility platforms and sidewalk robots. To address the long-tail problem, we plan to explore synthetic data generation via 4D Gaussian Splatting to create dynamic VRU "avatars" for augmenting rare events. We are committed to continuously extending the dataset and, by open-sourcing our platform, invite the community to contribute their own data and build upon our work.
\section{Conclusion}
\label{sec:conclusion}

In this work, we addressed the challenge of VRU-focused human action recognition in complex urban environments by introducing a complete pipeline from sensor platform to data acquisition and curation, to evaluation. We presented the \textit{FUSE-Bike}, a novel, hardware-synchronized multimodal perception bicycle designed to capture high-fidelity data from a VRU's perspective, making it uniquely suited for research in both AD and mobile robotics. With benchmarking analysis on \textit{BikeActions} dataset, we show that skeleton-based action classification can learn VRU actions, including hand gestures. Even though we only benchmark skeleton-based action classification, our dataset can also be used for estimated pose-based and video-based action classification. We plan to extend our dataset with more long-tail actions and explore the advantages of VRU action classification in downstream tasks such as behavior recognition, intention prediction, motion forecasting, and safe ego trajectory planning.

\section*{Acknowledgment}
This research was conducted within the project “Solutions and Technologies for Automated Driving in Town: An Urban Mobility Project”, funded by the Federal Ministry for Economic Affairs and Climate Action (BMWK), based on a decision by the German Bundestag, grant no. 19A22006N. 

\input{bib/bib-short.def}
\bibliographystyle{splncs04}
\bibliography{IEEEabrv, bib/mainBib}

\end{document}